\documentclass[sigconf,anonymous=False]{acmart}

\AtBeginDocument{%
  \providecommand\BibTeX{{%
    \normalfont B\kern-0.5em{\scshape i\kern-0.25em b}\kern-0.8em\TeX}}}
    
\usepackage{graphicx}
\usepackage{subcaption}
\usepackage{wrapfig}
\usepackage{dsfont}
\usepackage{textgreek}
\usepackage{scalerel}
\usepackage{subcaption}
\usepackage{bm}

\copyrightyear{2022} 
\acmYear{2022} 
\setcopyright{none}
\acmConference[Preprint]{}{version}{currently under review.}

\begin{document}


\title{RESHAPE: Explaining Accounting Anomalies in Financial Statement Audits by enhancing SHapley Additive exPlanations}

\author{Ricardo Müller}
\affiliation{%
    \institution{University of St\@.Gallen (HSG)}
    \city{St. Gallen}
    \country{Switzerland}
}
\email{ricardo.mueller@student.unisg.ch}

\author{Marco Schreyer}
\affiliation{%
    \institution{University of St\@.Gallen (HSG)}
    \city{St. Gallen}
    \country{Switzerland}
}
\email{marco.schreyer@unisg.ch}

\author{Timur Sattarov}
\affiliation{%
    \institution{Deutsche Bundesbank}
    \city{Frankfurt am Main}
    \country{Germany}
}
\email{timur.sattarov@bundesbank.de}

\author{Damian Borth}
\affiliation{%
    \institution{University of St\@.Gallen (HSG)}
    \city{St. Gallen}
    \country{Switzerland}
}
\email{damian.borth@unisg.ch}


\newcommand{\Tau}{\mathrm{T}}

\renewcommand{\shortauthors}{Müller, et al.}

\begin{abstract}

Detecting accounting anomalies is a recurrent challenge in financial statement audits. Recently, novel methods derived from Deep-Learning (DL) have been proposed to audit the large volumes of a statement's underlying accounting records. However, due to their vast number of parameters, such models exhibit the drawback of being inherently opaque. At the same time, the concealing of a model's inner workings often hinders its real-world application. This observation holds particularly true in financial audits since auditors must reasonably explain and justify their audit decisions. Nowadays, various Explainable AI (XAI) techniques have been proposed to address this challenge, e.g., SHapley Additive exPlanations (SHAP). However, in unsupervised DL as often applied in financial audits, these methods explain the model output at the level of encoded variables. As a result, the explanations of Autoencoder Neural Networks (AENNs) are often hard to comprehend by human auditors. To mitigate this drawback, we propose \textit{Reconstruction Error SHapley Additive exPlanations Extension} (RESHAPE), which explains the model output on an aggregated attribute-level. In addition, we introduce an evaluation framework to compare the versatility of XAI methods in auditing. Our experimental results show empirical evidence that RESHAPE results in versatile explanations compared to state-of-the-art baselines. We envision such attribute-level explanations as a necessary next step in the adoption of unsupervised DL techniques in financial auditing.

\end{abstract}

\ccsdesc[300]{Computing methodologies~Machine learning}

\ccsdesc[300]{Information systems~Enterprise resource planning}

\keywords{explainable artificial intelligence, artificial neural networks, audit, computer assisted audit techniques, accounting information systems, enterprise resource planning systems}

\maketitle

\section{Introduction}
\label{sec:introduction}

\begin{figure}[t]
    \center
  \includegraphics[width=\linewidth]{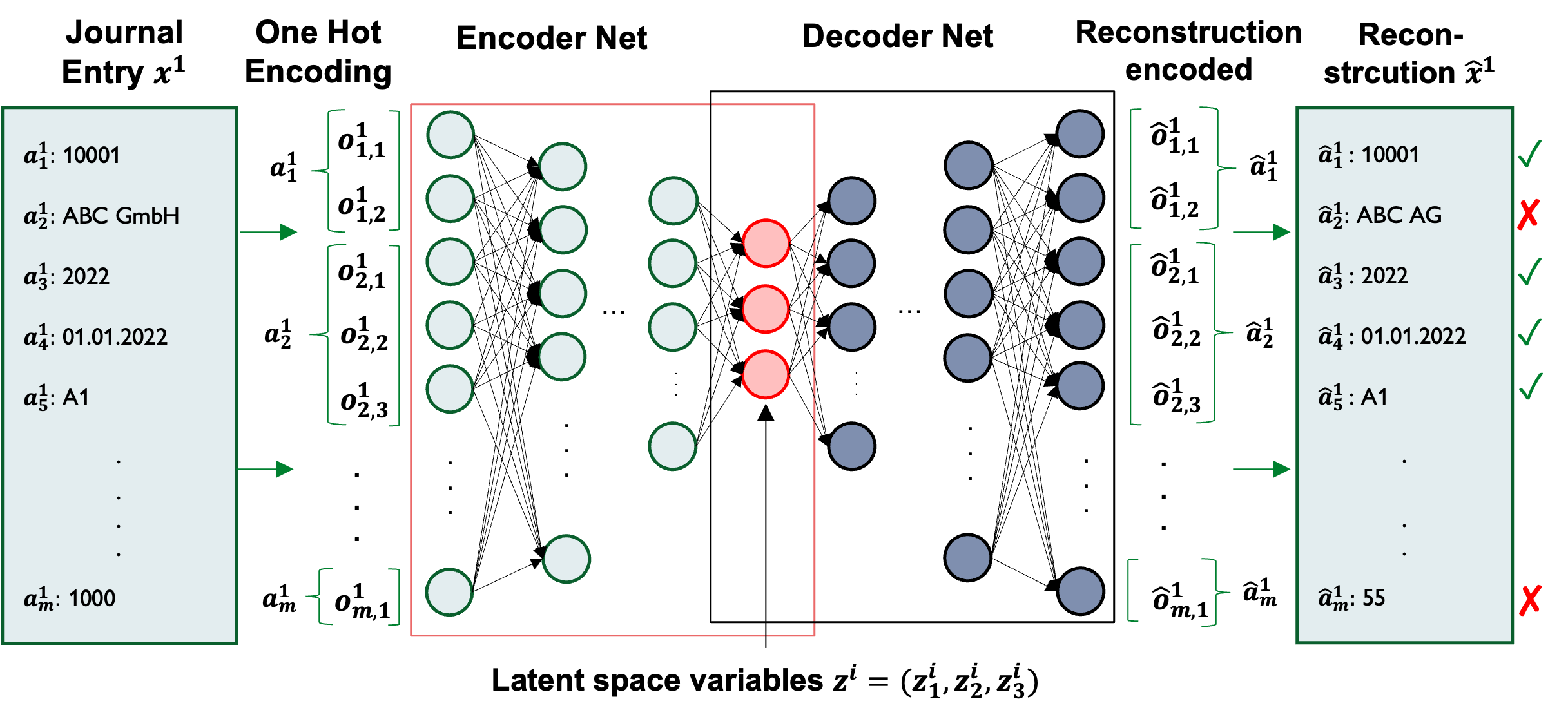}
  \caption[Schematic view of an autoencoder neural network]{Schematic view of an Autoencoder Neural Network (AENN) \cite{hinton2006} to reconstruct accounting records. The AENN architecture comprises the Encoder (green), the Bottleneck (red) and the Decoder (blue).}
  \label{fig:semantic_autoencoder}
  \vspace{-0.5cm}
\end{figure}

The financial statements of every major company are examined annually by an external auditor. During an audit, the auditor forms an opinion on whether the financial statements are free of material misstatements arising from fraud or error \cite{ISA200}. This protects both the shareholders and the economy from the negative fall-out that can result from flawed financial accounting. 


Accounting technology has evolved greatly over the past few decades. Nowadays, modern 
Enterprise-Resource-Planning (ERP) systems offer the possibility to record detailed information about every business transaction. This has significantly increased the amount of data available to the auditor. At the same time, the International Standards on Auditing (ISA) require that auditors perform an audit of the detailed accounting records, also referred to as \textit{Journal Entries} (JEs), to find indications of misstatements due to fraud or error (ISA 240 \cite{ISA240}, SAS 99 \cite{SAS99}). As a result of the increasing amount of available electronic audit evidence, computer-assisted audit techniques (CAATs) have emerged. CAATs include various techniques that allow the auditors to analyze the entire population of JEs. The currently used CAATs often utilize statistical methods to analyze unique attributes of a journal entry dataset at a time, limiting their anomaly detection performance \cite{coderre2012}. 

Driven by the steady progress in \textit{Artificial Intelligence} (AI), methods derived from \textit{Deep-Learning} (DL) \cite{lecun2015} have emerged into financial audits \cite{sun2019}. Recently, the effectiveness of using deep autoencoder neural networks (AENNs) to detect anomalies in large-scale JE datasets was successfully demonstrated \cite{schreyer2017, schultz2020}. However, the application of such highly parameterized models comes at the cost of model interpretability. As a result, such artificial neural networks are often perceived as opaque or 'black boxes' due to their vast amounts of underlying parameters. At the same time, model explainability is of paramount importance for auditors, as they are subject to extensive documentation and justification obligations\footnote{\textit{'The auditor shall prepare audit documentation that is sufficient to enable an experienced auditor, having no previous connection with the audit, to understand:[...] the nature, timing and extent of the audit procedures performed [...] The results of the audit procedures performed, and the audit evidence obtained; and significant matters arising during the audit, and the conclusions reached thereon.'} (ISA 230 \cite{ISA230})}. Ultimately, auditors face an inherent performance vs. interpretability trade-off when applying DL methods.

To derive explainable neural network outputs and alleviate their application in sensitive areas, model agnostic post-hoc explanation methods such as SHapley Additive exPlanations (SHAP) have been developed \cite{guidotti2018}. Lately, enhancements of SHAP have been applied in the context of AENN-based anomaly detection. Thereby, detected anomalies are explained either at a data \textit{instance-level}, such as the entire reconstructed JE (\textit{LossSHAP}) \cite{takeishi2019, roshan2021} or data \textit{encoding-level}, such as detailed reconstructed encodings (\textit{A-SHAP}) \cite{antwarg2019}.

In the audit context, \textit{instance-level} explanation methods often lack an explanation of the attributes that deem a JE to be anomalous. Furthermore, \textit{encoding-level} explanation methods often create a vast number of detailed explanations which scatter across multiple encodings. For auditors, interpreting both types of explanations is often challenging and time-consuming. In this work, we propose RESHAPE to explain accounting anomalies detected by AENNs on an aggregated \textit{attribute level}. In summary, we present the following contributions:

\begin{itemize}

    \item Introduction of a novel method of generating explanations for AENN that allows for \textit{attribute-level} explanations.
    \item Introduction of a comprehensive evaluation framework to benchmark XAI methods in a financial audit setup. 
    \item An extensive evaluation of RESHAPE against baselines using two synthetic and a real-world dataset. 
    
\end{itemize}


\noindent The remainder of this work is structured as follows: In section \ref{sec:relatedwork}, we provide an overview of related work. Section \ref{sec:methodology} follows with a description of the proposed methodology to explain detected accounting JEs. In section \ref{sec:evaluation_framework}, the introduced evaluation framework to benchmark XAI methods in a financial audit setup is presented. The experimental setup and results are outlined in section \ref{sec:experimental_setup} and section \ref{sec:experimental_results}. In section \ref{sec:summary}, the paper concludes with a summary. A reference implementation of the proposed methodology will be made available via [\textit{url redacted due to double-blind review}]. 




\section{Related Work}
\label{sec:relatedwork}

The subsequent literature review provides an overview of the relevant related work focusing on: (i) the detection of accounting anomalies using AENNs, (ii) the application of XAI methods for anomaly detection, and (iii) the evaluation of XAI methods.

\subsection{Detection of Accounting Anomalies} 

\noindent Surveys on anomaly detection methods were conducted by \cite{Chandola09, ahmed2016, kwon2019, chalapathy2019}. Since Hawkins et al. first introduced the application of AENNS to detect data anomalies \cite{hawkins2002, williams2002, goodfellow2016}, their capabilities have been demonstrated on different datasets. In the context of financial fraud, AENNs have been used to detect anomalies in credit card transaction datasets \cite{kazemi2017, pumsirirat2018}. However, fewer publications attempt to employ AENNs in the context of auditing. The application of AENNs to detect anomalies in large-scale accounting data presented by Schreyer et al. \cite{schreyer2017} is closely related to this paper. In a subsequent work, the authors \cite{schreyer2019} demonstrated that the JE representation learned by the AENN can be used to identify anomalies. Building on this work, Schultz et al. \cite{schultz2020} compared the results of an AENN-based audit of JEs to the audit findings of human auditors. Furthermore, Zupan et al. \cite{zupan2020} used Long Short-Term Memory AAENs to detect temporal anomalies in journal entry data. In addition, Nonnenmacher et al. \cite{nonnenmacher2021} and Schreyer et al. \cite{schreyer2020l} demonstrated that AENNs could be used to improve audit sampling during an audit process. Furthermore, it was shown that AENNs can be trained in a self-supervised learning setup to detect accounting anomalies and complete additional down-stream audit tasks \cite{schreyer2021}.

\subsection{Explainable AI in Anomaly Detection} 

\noindent An overview of methods to explain anomalies was presented by Yepamo et al. in \cite{yepmo2022}. As most commonly used anomaly detection methods offer limited intrinsic explainability, post-hoc XAI methods are preferred \cite{yepmo2022}. Two popular XAI methods for generating local and post-hoc explanations \textit{Local Interpretable Model-Agnostic Explanations} (LIME)
and \text{SHapley Additive exPlanations} (SHAP). The LIME method \cite{ribeiro2016} generates explanations by approximating the models' local behavior using a linear, and thus interpretable, regression. In contrast, the SHAP method \cite{lundberg2017}, draws from the Shapley values, a renowned game theory concept, to generate explanations. In \cite{takeishi2019} Takeishi used SHAP to explain the anomaly score of anomalies detected using a Principal Component Analysis (PCA). Previous publications in this domain focused on frameworks to generate visual explanations \cite{liu2017, collaris2018, goodall2018}. Roshan \& Zafar \cite{roshan2021} recently used the same method to explain computer network anomalies detected by an AENN. In a premier attempt to apply XAI methods to generate explanations for AENN outputs, Antwarg et al. \cite{antwarg2019} compared the performance of both SHAP and LIME and found that SHAP generates superior explanations. Additional publications demonstrated the usefulness of combining SHAP and AENNs \citep{chawla2020, giurgiu2019, psychoula2021}. Recently, two publications focused on applying XAI in the financial audit domain were published. Rebstadt et al. \cite{rebstadt2022} developed a role model for XAI in auditing. Furthermore, Gnoss et al. \cite{gnoss2022} performed a user study to determine the usefulness of XAI in the audit domain. However, they resorted to explaining a surrogate neural network trained on the dataset labelled by the AENN, as generating global explanations using the AENN directly is particularly challenging.

\subsection{Explainable AI Evaluation Frameworks} 

The evaluation of XAI methods is a rather broadly defined goal and covers a large set of methods, processes, and metrics. Surveys about the evaluation of XAI methods were published by \cite{doshi2017, zhou2021}. Defining appropriate and generally applicable metrics to evaluate the quality of explanations \cite{nguyen2020, alvarez2018, velmurugan2021, tritscher2020, visani2020}, the field is actively researched and hasn't converged towards a set of standard metrics. Nguyen \& Martínez \cite{nguyen2020} even argue that a general computational benchmark is unlikely to be possible, as the required qualities of explanations heavily depend on the context. Nevertheless, some researchers made first advances to integrate different metrics into comprehensive benchmarks \cite{yang2019,amparore2021, liu2021}.  Yang and Kim \cite{yang2019} introduced the Benchmark Interpretability Methods framework (BIM), which offers a set of tools to quantitatively compare a model’s ground truth to the output of interpretability methods. Amparore et al. \cite{amparore2021} released a library that offers a comprehensive set of evaluation metrics to compare evaluation methods. These metrics include conciseness, local fidelity, local concordance, reiteration similarity and prescriptivity. Liu et al. \cite{liu2021} released a library to evaluate explainability methods using a synthetic dataset to develop a ground truth. This allows an exact computation of the popular metrics faithfulness, monotonicity, remove and retrain (ROAR), ground truth Shapley values and infidelity.

\vspace{0.5em}

The above literature survey demonstrated that whilst AENNs have often been used to detect anomalies in financial datasets, publications on explaining AENNs used in the financial audit domain are scarce. Multiple methods of applying SHAP to generate explainable AENNs have been proposed but have not yet been compared. Additionally, the proposed methods only allow for data instance-level or encoding-level explanations. Finally, benchmarking the different XAI approaches is challenging due to the absence of a standardized framework of evaluation metrics. 

\section{Methodology}
\label{sec:methodology}

This section describes the distinct steps of the proposed \textit{Reconstruction Error SHapley Additive exPlanations Extension} (RESHAPE). Furthermore, it provides details on the application background and baseline methods.

\vspace{-0.2cm}


\subsection{Accounting Journal Entries}
\label{sec:accounting_journal_entries}

Formally, a journal entry dataset $X$ comprises $N$ JEs \scalebox{0.85}{$X = \{x^1,x^2,...,x^n\}$}. Each JE \scalebox{0.85}{$x^i$} holds a tuple of attribute values, \scalebox{0.85}{$x^i = \{a_1^i,a_2^i,...,a_M^i,a_1^i, a_2^i...a_K^i\}$} with $j=1,2,3...M$ categorical attributes and $l=1,2,3,...K$ numerical attributes. In total, a JE consists of $T=M+K$ attributes. The general objective of the AENNs used in this work is to detect accounting anomalies. Similar to Breunig et al. \cite{breunig2000} we distinguish two classes of anomalies: i) JEs that exhibit exceptional attribute values (\textit{global anomalies}) and ii) JEs that exhibit an unusual combination of attribute values (\textit{local anomalies}). We enhance the classification by introducing global $TypeA_k$ and local $TypeB_k$ anomalies to measure a JE's' anomalousness' level. \textbf{\bm{$TypeA_k$} Anomalies} define JEs where $k$ attributes contain values that are unprecedented in a given dataset. For example, if no JE with a Posting Key of `D18' could be observed, then an $TypeA_1$ anomaly can be created by setting the \textit{Posting Key} attribute value of a randomly sampled JE to `D18'. Ultimately, a JE that is classified as $TypeA_k$ anomaly exhibits \textit{$k$} attribute values that are not observable in other JEs of the dataset. In real-world accounting data, $TypeA_k$ anomalies are often caused by rarely used document types, accounts, users, etc. Hence, $TypeA_k$ anomalies are associated with a high risk of error. \textbf{Local \bm{$TypeB_k$} Anomalies} define JEs, where $k$ attributes contain values that are observable in a given dataset. However, a $TypeB_k$ anomaly comprises an anomalous combination of the top $n$ most observable attribute values. For example, if a JE differs from each JE in the dataset by at least e.g., three attributes, it can be considered a $TypeB_3$ anomaly. In real-world accounting data, $TypeB_k$ anomalies can stem from unusual combinations, e.g., posting time and user accounts. Hence, they are generally associated with high fraud risk. In a real-world audit setup, auditors aim to detect both classes of anomalies. 

\subsection{Autoencoder Neural Networks (AENNs)} 
\label{sec:autoencoder_neural_networks}

An \textit{Autoencoder Neural Network} (AENN), as illustrated in Fig. \ref{fig:semantic_autoencoder} defines a feed-forward multi-layer neural network that is trained to reconstruct its input \cite{hinton2006}. In general, an AENN consists of two non-linear functions, namely an \textit{Encoder Network} $f_\theta(x^i)$ and a \textit{Decoder Network} \scalebox{0.90}{$g_\psi(z^i)$} \cite{baldi2012}. The encoder function \scalebox{0.90}{$z^i=f_\theta(x^i)$} maps the input \scalebox{0.90}{$x^i \in \mathbb{R}^{n}$} to a latent space representation \scalebox{0.90}{$z^i \in	\mathbb{R}^{p} $} in the latent space $Z$. The decoder network \scalebox{0.90}{$\hat{x}^i = g_{\psi}(z^i)$}  maps the latent space representation $z^i$ to the model output \scalebox{0.90}{$\hat{x}^i \in	\mathbb{R}^{n}$}. Formally, the learning objective of an AENN is to minimize the difference between the input JE $x^i$ and its reconstruction $\hat{x}^i$.

\begin{equation}
    \arg \min_{\psi\theta} || x^i - g_\psi(f_\theta(x^i)) ||
    \label{eq:autoencoder}
\end{equation} 

\noindent where $f_\theta(\cdot)$ denotes the encoder function with its parameters $\theta$. $g_\psi(\cdot)$ denotes the decoder function with corresponding parameters $\psi$. The input is described by $x^i$. To ensure that the AENN learns a rich data representation, the dimension of the bottleneck layer $p$ is decreased compared to the input and output layer dimension $n$. The constraint of $dim(p)<dim(n)$ forces the AENN to learn latent representations $z^i$ of the most salient input data features of the input data \cite{goodfellow2016}. Furthermore, the architectural structure necessitates that the AENN draws from patterns in the data that generalize for the majority of the JEs in the dataset. A JE can't be faithfully reconstructed if it does not correspond to the general patterns learned by AENN during the unsupervised learning procedure \cite{goodfellow2016, baldi2012}. Thereby, the AENNs \textit{Reconstruction Error} (RE), denoted as $\mathcal{L}(x^i; \hat{x}^i)$, quantifies the difference between an original input JE $x^i$ and its reconstruction $\hat{x}^i$. Upon successful model training, the RE is then used to quantify the degree of `anomalousness' of a given JE. If the RE of a journal \scalebox{0.9}{$\mathcal{L}(x^i;\hat{x}^i)$} exceeds a threshold $\delta$, the entry corresponding to \scalebox{0.9}{$\mathcal{L}(x^i;\hat{x}^i)> \delta$}, is flagged as an anomaly, denoted as $\Tilde{x}^i$ in the following.

\begin{figure*}[t]
  \includegraphics[width=\linewidth]{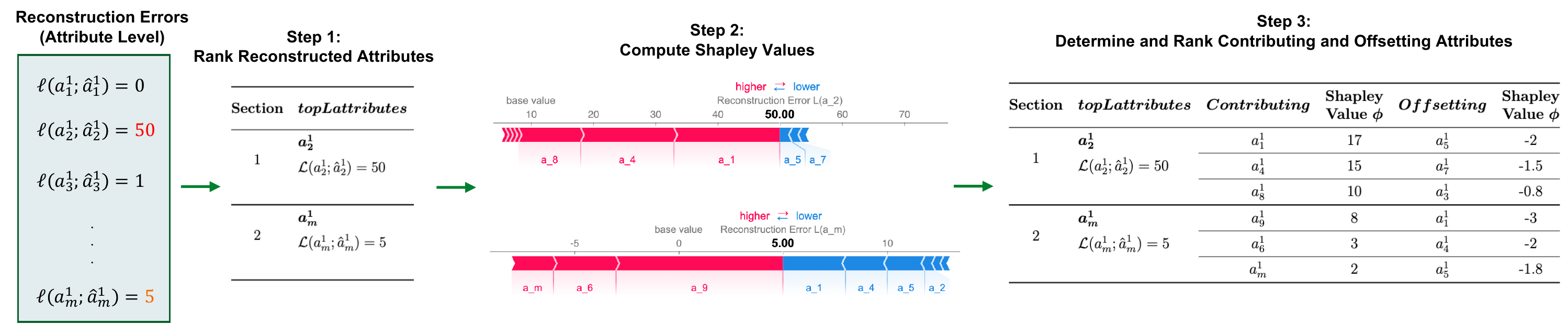}
  \caption{Process of explaining an anomalous instance using RESHAPE. Firstly, the JE is reconstructed by the AENN, and the attribute-based reconstruction errors are used to sort the attributes, generating the $topLattributes$ list. For each attribute in $topLattributes$, an explanation is generated using SHAP. The explanation quantifies the influence of all attribute values on the reconstruction error of the attribute in question, for example $\mathcal{L}(a^1_2;\hat{a}^1_2$). The resulting Shapley values are then used to split the attributes based on whether they contributed or offset the reconstruction error and are presented in tabular form.}
  \label{fig:RESHAPE_Overview}
\end{figure*}

\subsection{SHapley Additive exPlanations (SHAP)}
\label{sec:shapeley_additive_explanations}

Lundberg \& Lee in \citep{lundberg2017} proposed, the \textit{SHapley Additive exPlanations} (SHAP), combining existing XAI methods and \textit{Shapley Values} \citep{shapley1951}. Shapley Values, a concept derived from game theory, measure the individual payoff contribution by a single player $j$ in a multi-player game of $K$ players. Therefore, Shapley Values can be used as a basis for a fair distribution $\upsilon(K)$ of the payoff among the $K$ players. Formally, the Shapely Value of a particular player $j$ is calculated according to:

\begin{equation}
    \phi_j= \sum\nolimits_{S\subseteq K\setminus_j}\frac{\vert S \vert !\times(\vert K \vert - \vert S \vert - 1)!}{K!}(\upsilon(S\cup j) -\upsilon(S)), 
    \label{eq:shapley}
    \vspace{0.2cm}
\end{equation}

\noindent where $\phi_j$ denotes the Shapley Value of player $j$, the set of all players is denoted by $K$. Furthermore, $S\subseteq K\setminus_j$ denotes the set of all possible subsets of players excluding the $j$-th player. Ultimately, the payoff function $\upsilon(\cdot)$ defines the (hypothetical) payoff that a subset of players achieves. With SHAP, Lundberg \& Lee adopted the Shapley Value concept to explain the behavior of ML models. Thereby, explanations $e(x^i)$ are crafted considering an ML model as a game. Using this analogy, the output of a model is viewed as the game's payoff. The main idea of SHAP is to explain ML models based on the importance of an individual data feature for the model output. The importance of a particular feature $j$ is determined by its Shapely Value $\phi_j$. However, SHAP is designed to only explain a single model output while treating the model to be explained as a `black box' \cite{lundberg2017}. This is a major limitation in unsupervised DL learning, particularly AENN-based anomaly detection. AENNs are typically designed to output a vector of outputs rather than a single output. Recently, two methods have been proposed that adapt SHAP to facilitate the output structure of AENNs. Both methods, namely \textit{LossSHAP} \cite{takeishi2019} and \textit{A-SHAP} \cite{antwarg2019}, explain the detected anomalies at different levels of abstraction as described in the following.

\vspace{0.5em}

\noindent \textbf{Instance-Level SHAP (LossSHAP) \cite{takeishi2019}:} In the LossSHAP setup, the reconstruction error of entire data instances is explained. In the audit context, LossSHAP creates explanations at the \textit{transaction-level} of entire JEs. Thereby, a single explanation is generated that aggregates information of multiple JE attributes. However, such an explanation often lacks a sufficient level of detail to be examined by a human auditor.

\vspace{0.5em}

\noindent \textbf{Encoding-Level SHAP (A-SHAP) \cite{antwarg2019}:} In the A-SHAP setup, the reconstructions of individual encodings are explained. In the audit context, A-SHAP creates explanations at the detailed \textit{encoding-level} of attribute encodings. Thereby, multiple explanations are generated, each corresponding to a single JE encoding. However, such granular explanations are often difficult to comprehend by a human auditor. 

\vspace{0.5em}

\noindent To facilitate explanations on an attribute level using SHAP, we propose the \textit{Reconstruction Error SHapley Additive exPlanations Extension} (RESHAPE). Using RESHAPE allows auditors to leverage an attribute-level explanation structure that combines the advantages of the LossSHAP and the A-SHAP method.

\subsection{Reconstruction Error SHapley Additive exPlanations Extension (RESHAPE)}
\label{sec:reconstruction_error_shapley_additive_explanations_extension}

To introduce RESHAPE, let \scalebox{0.90}{$o_{m} = \{o_{1,m}, o_{2,m},...,o_{k,m}\}$} define the encoding of a particular JE attribute $a_{m}$, where $k$ denotes the number of encoding dimensions of the $m$-th attribute. Furthermore let \scalebox{0.90}{$\hat{o}_{m} = \{\hat{o}_{1,m}, \hat{o}_{2,m},...,\hat{o}_{k,m}\}$} define encodings of the reconstructed JE attribute $\hat{a}_{m}$ derived by the AENN. To establish the RESHAPE approach for each attribute $a_{m}$ and its reconstruction $\hat{a}_{m}$, we define an attribute loss, as formally defined by:

\begin{gather}
    \mathcal{L}(a_m;\hat{a}_m) = \sum\nolimits_{k=1}^{K}\mathcal{L}(o_{m,k};\hat{o}_{m,k}),
    \label{eq:summed_error}
\end{gather}

\noindent where $a_m$ denotes the $m$-th attribute of a JE and $o_{m,k}$ denotes its $k$-th encoded attribute dimension. In the following, the three RESHAPE steps performed to create an attribute-level explanation of an anomalous JE are introduced. The distinct process steps are illustrated in Fig. \ref{fig:RESHAPE_Overview}.

\vspace{0.5em}

\noindent\textbf{Step 1: Rank Reconstructed Attributes}: In a first step, the JE's attributes are ranked according to their reconstruction errors \scalebox{0.90}{$\mathcal{L}(a_m^i;\hat{a}_m^i)$}. This process results in a ranked list of attributes \scalebox{0.90}{$\{a_{(1)}, a_{(2)} \cdots a_{(m)}\}$}, where \scalebox{0.90}{$\mathcal{L}(a^i_{(1)};\hat{a}^i_{(1)}) \geq \mathcal{L}(a^i_{(2)};\hat{a}^i_{(2)})$} \scalebox{0.90}{$\geq$} \scalebox{0.90}{$\cdots$} \scalebox{0.90}{$\geq$} \scalebox{0.90}{$\mathcal{L}(a^i_{(m)};\hat{a}^i_{(m)})$}. The top $l$ ranked attributes, exhibiting the highest reconstruction errors, are selected, formally defined by \scalebox{0.90}{$topLattributes =$} \scalebox{0.90}{$\{a_{(1)}, a_{(2)}, \cdots, a_{(l)} \}$}.


\vspace{0.5em}

\noindent\textbf{Step 2: Compute Shapley Values}: In a second step, for each attribute in \scalebox{0.90}{$topLattributes$} a SHAP explanation run is conducted. The objective of the explanation run is to determine for each attribute value its impact on the attribute reconstruction error. Thereby, the reconstruction error of the attribute to be explained \scalebox{0.90}{$\upsilon(\cdot) = \mathcal{L}(a_l^i;\hat{a}_l^i)$} is used as the payoff function $\upsilon(\cdot)$ of the SHAP runs. Ultimately, the different runs result in a list of attribute attribution values, formally defined by \scalebox{0.90}{$\phi^l = \{\phi^l_1, \phi^l_2, \dots, \phi^l_m\}$}.

\vspace{0.5em}

\noindent\textbf{Step 3: Rank Contributing and Offsetting Attributes}: In a third step, the attribute attribution values \scalebox{0.9}{$\phi^l$} form the basis to determine an explanation's \textit{contributing} and \textit{offsetting} attributes. Attributes \scalebox{0.85}{$a_j^i$} exhibiting a positive (negative) error attribution \scalebox{0.85}{$\phi^l_j < 0$} (\scalebox{0.85}{$\phi^l_j > 0$}) are marked as contributing (offsetting) attributes. Finally, attributes exhibiting a high offsetting or contributing magnitude are considered as an explanation for the JE's anomalousness.

\vspace{0.5em}

\noindent Upon completion of the process steps, the generated explanations are presented to a human auditor. Based on the explanations, the auditor can deduce on an attribute-level why the AENN flagged a particular as an anomaly.

\section{Evaluation Framework}
\label{sec:evaluation_framework}

In the following, we propose a set of evaluation measures to evaluate the quality of explanations in the financial audit setting. The framework evaluates three desirable characteristics that explanations should encompass to enable auditors to derive actionable decisions, namely (i) \textit{fidelity}, (ii) \textit{stability}, and (iii) \textit{robustness}. 

\subsection{Explanation Fidelity} 
\label{sec:explanation_fidelity}

The fidelity of an explanation is defined as the explanation's ability to capture the underlying local dynamics of the model \cite{yeh2019}. This is particularly important in the audit context, as false decisions based on inaccurate explanations could have severe consequences. We evaluate the fidelity of generated explanations using three metrics. 

\subsubsection{Mean Reciprocal Rank ($MRR_r$) Measure \cite{tritscher2020, antwarg2019}:} This metric measures the average rank of known attributes that contributed to the anomalousness of a JE. Thereby, the $MRR_r$ measures the average inverse rank of an explanation's first relevant attribute, as formally defined by:

\begin{gather}
    MRR_r = \frac{1}{M}\sum\nolimits_{i=1}^M\frac{1}{rank_R^i},
    \label{eq:MRR}
\end{gather}
    
\noindent where $M$ denotes the number of explanations, \scalebox{0.85}{$R=\{a_j^i,...,a_r^i\}$} the set of explanation relevant attributes. Thereby, the variable \scalebox{0.90}{$rank_R^i$} denotes the first occurrence of a relevant attribute in the explanation. 

\subsubsection{Hits@n Measure \cite{takeishi2019}:} This metric determines if a relevant attribute resides among the top-$n$ ranked attributes of an explanation. This is an important property, as auditors might focus on most explanatory attributes when interpreting a detected anomalous JE. Given an explanation, the $Hits@n^i$ measure is formally defined as:

\begin{gather}
    Hits@n^i = 
    \begin{cases}
        1, & \text{if } rank_R^i \leq n \\
        0, & \text{otherwise}
    \end{cases}
    \label{eq:Hits}
\end{gather}
    
\noindent where $n$ denotes the rank up to which attributes in the set of explaining attributes \scalebox{0.85}{$R=\{a_j^i,...,a_r^i\}$} are considered. Furthermore, the variable \scalebox{0.90}{$rank_R^i$} denotes the first occurrence of a relevant attribute in an explanation.

\subsubsection{Reduction of Anomaly Score:} This metric measures how an anomalous JE's attribute needs to be adjusted to convert it into a regular JE. The objective is thereby to constantly minimize the reconstruction error $\min(\mathcal{L}(x^{'i};\hat{x}^{'i})_n)$ by be sequentially adjusting attribute values of the most explanatory attributes \scalebox{0.85}{$R=\{a_j^i,...,a_r^i\}$}. The reconstruction error decrease is then determined according to:

\begin{gather}
 error\%_n^i = \frac{\mathcal{L}(x^{'i};\hat{x}^{'i})_n}{\mathcal{L}(x^i;\hat{x}^i)},
    \label{eq:error_percent}
\end{gather}

\noindent where \scalebox{0.90}{$\mathcal{L}(x^{'i};\hat{x}^{'i})_n$} denotes the reconstruction error upon conversion of the top-$n$ explanatory attributes and \scalebox{0.90}{$\mathcal{L}(x^i;\hat{x}^i)$} the original reconstruction error of the JE.



\subsection{Explanation Stability} 
\label{sec:explanation_stability}

In the audit context, supervisory bodies demand the reproducibility of analytical audit procedures. The explanation stability \cite{mishra2021} measures the variability of an explanation over different explanation run parameters. For each explanation $K$ runs are conducted, and the $n$ top-ranked explanatory attributes are collected. Afterwards a \textit{Stability Index} is determined, formally defined as:

\begin{gather}
    S_n = \sqrt{\frac{V_{1} + V_{2} + ... + V_{i} + ... + V_{n}
    }{n}}
    \label{eq:instability_index}
\end{gather}

\noindent where \scalebox{0.90}{$V_{i}= Var(r_{1_i},...r_{k,i})$} denotes the variance in rank \scalebox{0.90}{$r_{k,i}$} of the $i$-th explanatory attribute in the $K$ explanation runs.

\subsection{Explanation Robustness} 
\label{sec:explanation_robustness}

The explanation robustness \cite{antwarg2019} measures an explanation's attribution to uninformative attributes. This is relevant in an audit setting, where auditors don't want to be misguided by irrelevant attributes. The robustness is determined by the MRR of uninformative attributes ($MRR_u$) as defined in Eq. \ref{eq:MRRU}.

\begin{gather}
    MRR_u = \frac{1}{M}\sum\nolimits_{i=1}^M\frac{1}{rank_U^i},
    \label{eq:MRRU}
\end{gather}

\noindent where $M$ denotes the number of explanations, \scalebox{0.85}{$U=\{a_j^i,...,a_r^i\}$} the set of explanation uninformative attributes. Thereby, the variable \scalebox{0.90}{$rank_U^i$} denotes the first occurrence of an uninformative attribute in the explanation. 


\section{Experimental Setup}
\label{sec:experimental_setup}

In this section, we describe the setup of our experiments to evaluate the proposed RESHAPE method. Our evaluation follows a two-step approach: First, a variety of AENN models are trained based on deliberately selected datasets. Second, the AENN models, particularly the detected anomalies, are explained using RESHAPE and baseline XAI methods. The explanations are evaluated using the metrics presented in section \ref{sec:evaluation_framework}.

\subsection{Datasets and Data Preparation}
\label{subsec:datasets}

We use two synthetic datasets and a publicly available real-world financial payment dataset to conduct a comprehensive evaluation. In the following, a summary of each dataset is presented:

\begin{itemize}

\item \textbf{Synthetic Boolean Dataset}: This dataset consists of 15 random boolean variables as well as 5 dependent variables implementing boolean functions such as the $OR$ or $XOR$ operator.\footnote{\scalebox{0.9}{url redacted due to blind review}} The dataset encompasses a total of 2.09 M records, including 75,000 inserted synthetic anomalies. The encoding resulted in a total of 20 encoded dimensions for each data instance $x^{i} \in \mathcal{R}^{20}$.

\item \textbf{Synthetic Accounting Dataset}: This dataset encompasses 533,091 transactions that mimic JEs extracted from SAP-ERP systems comprising $7$ categorical and two numerical attributes.\footnote{\scalebox{0.9}{\url{https://github.com/GitiHubi/deepAI}}} A total of 280 synthetic anomalies were inserted. The encoding resulted in a total of 704 encoded dimensions for each of the JE records $x^{i} \in \mathcal{R}^{704}$.

\item \textbf{Real-World Payments Dataset}: This real-world dataset encompass a total of 238,894 city payments comprised of 10 categorical and one numerical attribute.\footnote{\scalebox{0.9}{\url{https://www.phila.gov/2019-03-29-philadelphias-initial-release-of-city-payments-data/}}} A total of 300 synthetic anomalies were inserted. The encoding resulted in a total of 8,565 encoded dimensions for each of the city’s vendor payment record  $x^{i} \in \mathcal{R}^{8,565}$.

\end{itemize}

\noindent Both the `Synthetic Accounting Dataset' and the `Real-World Payments Dataset' exhibit a high similarity to JE data examined in financial statement audits, e.g., manual payments or payments runs.

\subsection{AENN Training Setup}

To train dataset-specific models, we use fully-connected layers exhibiting different architectural setups. Tab. \ref{tab:AENN_setup} provides an overview of the respective setups. In each layer we use Leaky-ReLU non-linear activations functions with scaling factor $\alpha = 0.4$ except for the encoder's and decoder's final layers. When optimizing the AENN models, we compute the binary cross entropy error of a given reconstructed JE $\hat{x}_{i}$, formally defined as:

\begin{gather}
    \mathcal{L}^{BCE}(x_{i};\hat{x}_{i}) = \frac{1}{N} \sum\nolimits_{i=1}^{N} x_{i} \cdot \log(\hat{x}_{i}) + (1 - x_{i}) \cdot \log(1 - \hat{x}_{i}),
    \label{eq:bce_error}
\end{gather}

\vspace{0.2cm}

\noindent where $x_{i}$ denotes the original encoded JE. We used a batch size of $\rho=$ 128 and Adam optimization with $\beta_{1}=0.9$, $\beta_{2}=0.999$ in all our experiments. The models are trained for a max. of 500 (Boolean Dataset), 5 (Accounting Dataset), and 25 (Payment Dataset) training epochs using a learning rate of 0.0001. We applied early stopping once the loss converged.

\begin{table}[htbp]
\centering
\caption{Overview of architectural details used to train the AENN models on the distinct datasets.}
\label{tab:AENN_setup}

\resizebox{0.45\textwidth}{!}{%
\begin{tabular}{@{}ll@{}}
Dataset           & Neurons per Fully-Connected Network Layer                                                                        \\\midrule
Synthetic Data                                                             & 20-18-16-15-16-18-20                                                                                \\
Accounting Data                                                           & 704-512-256-128-[$\cdots$]-8-4-3-4-8-[$\cdots$]-128-256-512-704                                                   \\
Payment Data                                                        & 6,358-5,096-2,048-[$\cdots$]-8-4-3-4-8-[$\cdots$]-2,048-5,096-6,358

\\\bottomrule

\end{tabular}%
}
\vspace{-0.3cm}
\end{table}

\begin{table}[htbp]
\centering
\caption{Overview of architectural details used to train the AENN models on the distinct datasets.}
\label{tab:AENN_setup}

\resizebox{0.25\textwidth}{!}{%
\begin{tabular}{@{}ll@{}}
Dataset           & Backgroundset Size $\eta$                                                                    \\\midrule
Synthetic Data                                                             & 500                                                                              \\
Accounting Data                                                           & 500                                                  \\
Payment Data                                                        & 250

\\\bottomrule

\end{tabular}%
}
\vspace{-0.3cm}
\end{table}

\subsection{Explanation Runs Setup}

Upon successful AENN model training, the anomalies are explained using RESHAPE and baseline XAI methods. We use a background set $\eta$ of 500 data instances for the boolean and accounting data in all our SHAP explanation runs. For the transactions dataset we use a reduced background set size  $\eta = 250$ due to the high number of encoded dimensions (see Tab. \ref{tab:AENN_setup}). Each explanation run was repeated three times using different random seed initialization. The experimental results for a \textit{Random} ordering of attributes are recorded As an additional baseline.


\begin{figure*}[!th]
    \begin{subfigure}{0.24\textwidth}
        \centering
        \includegraphics[width=\textwidth]{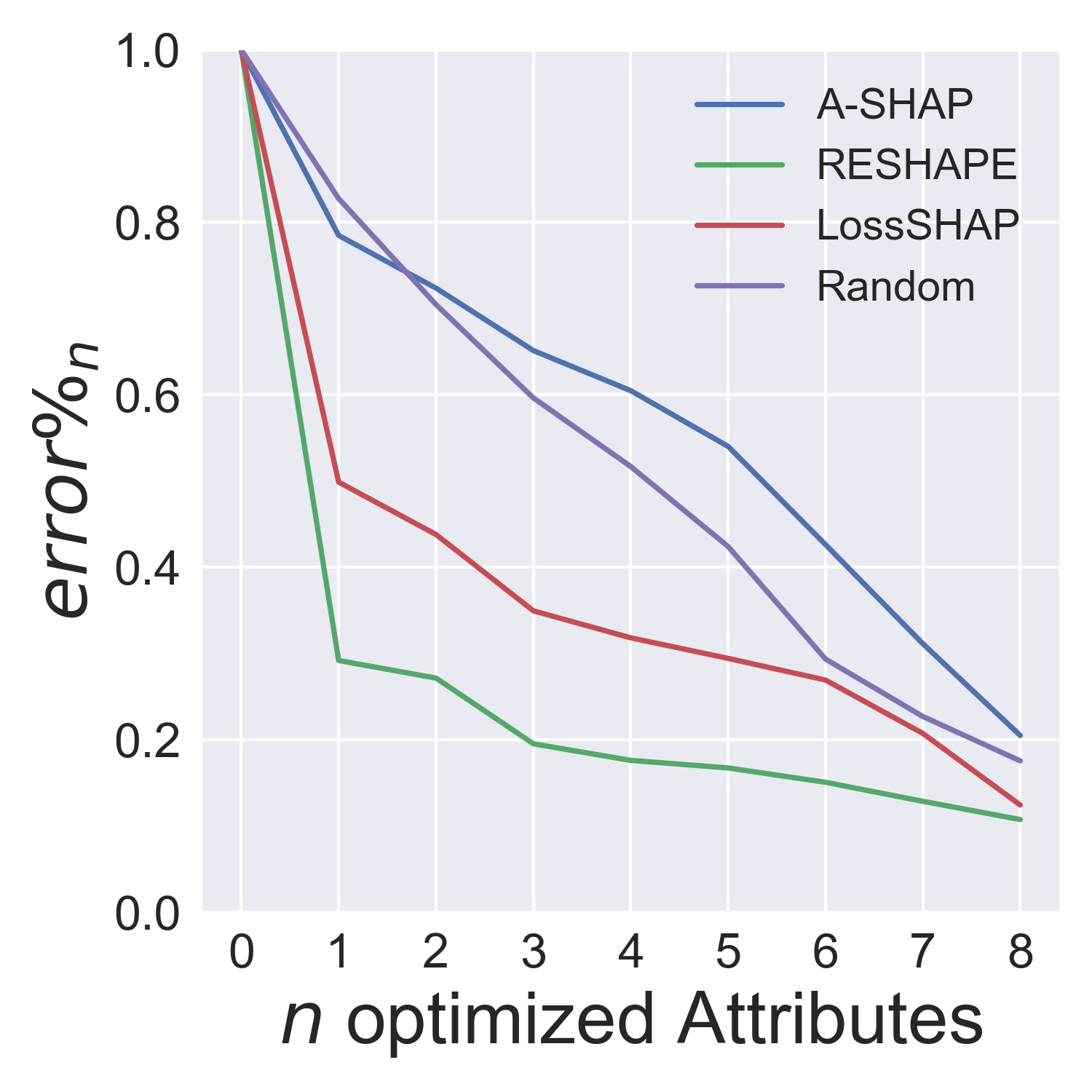}
        \caption{Accounting Data}
    \end{subfigure}
    \begin{subfigure}{0.24\textwidth}
        \centering
        \includegraphics[width=\textwidth]{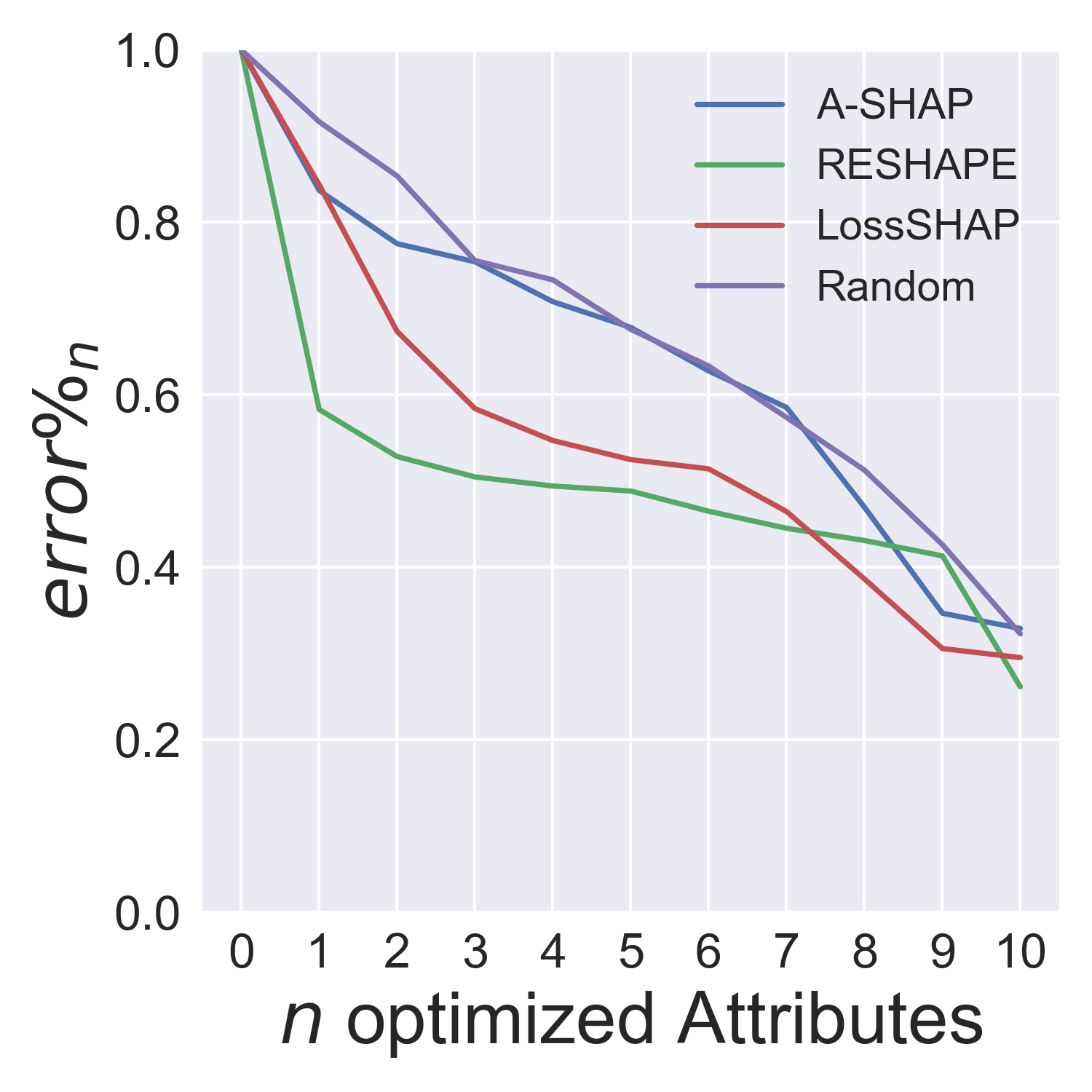}
        \caption{Payment Data}
    \end{subfigure}
    \begin{subfigure}{0.24\textwidth}
        \centering
        \includegraphics[width=\textwidth]{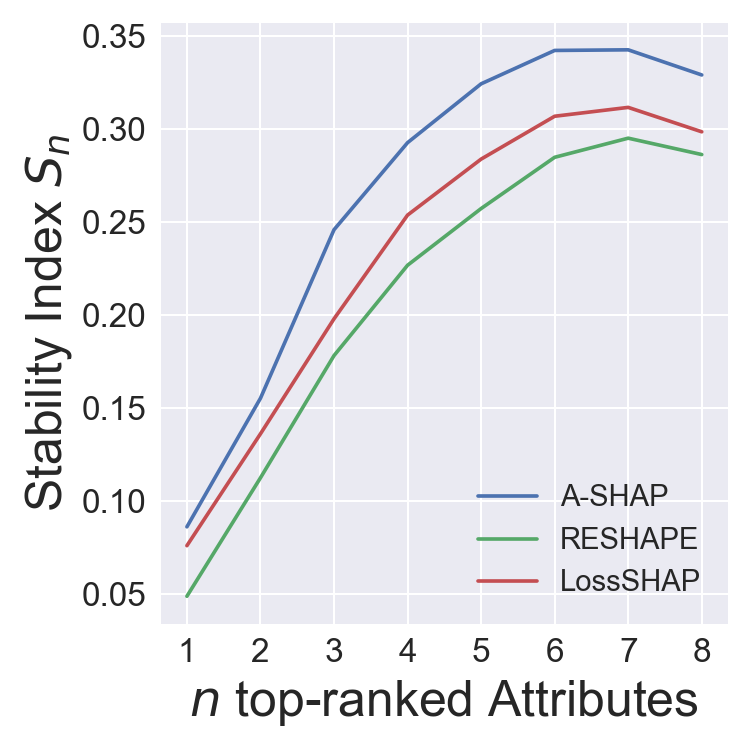}
        \caption{Accounting Data}
    \end{subfigure}
    \begin{subfigure}{0.24\textwidth}
        \centering
        \includegraphics[width=\textwidth]{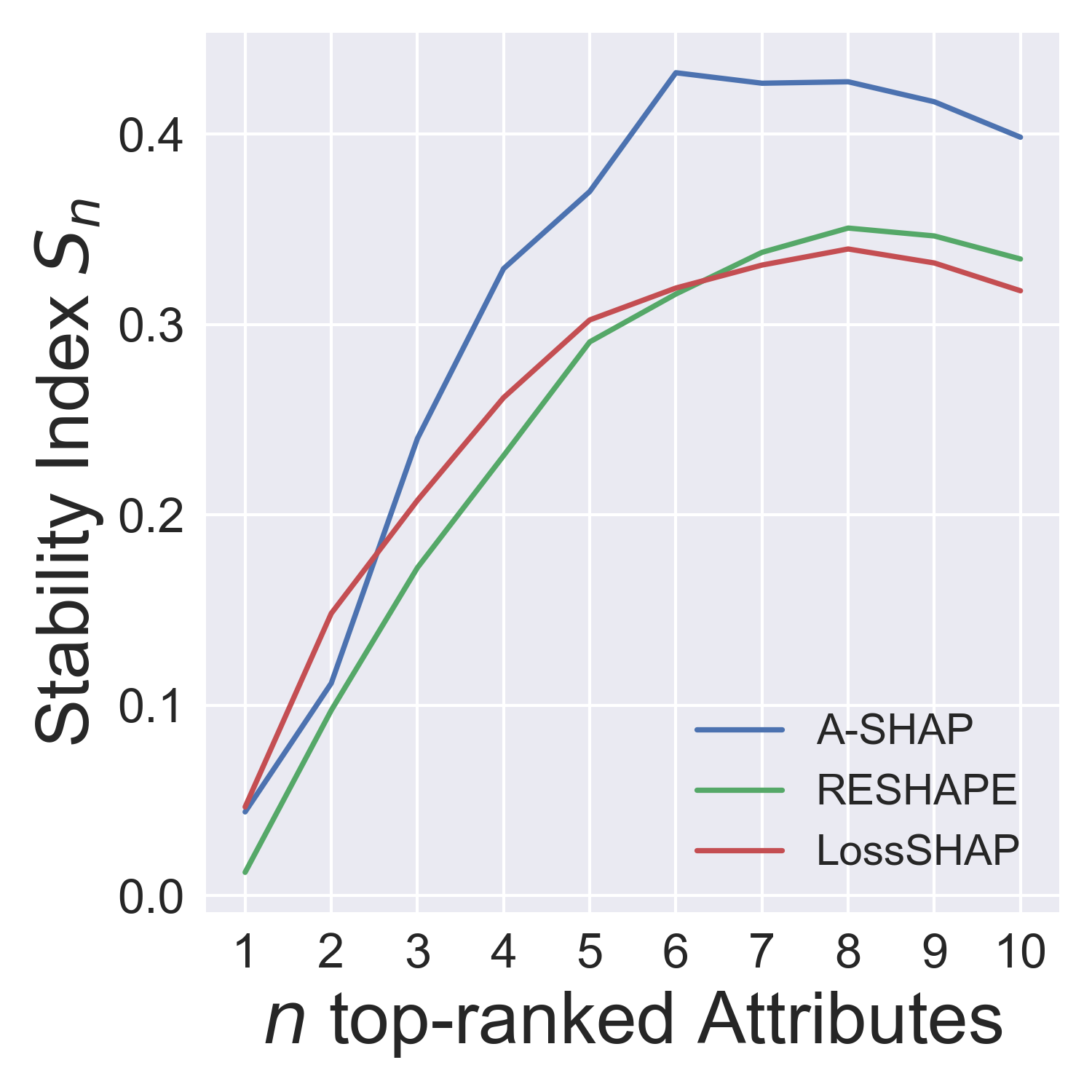}
        \caption{Payment Data}
    \end{subfigure}
    \vspace{-2mm}
    \caption{Results of reduction of anomaly score evaluation (a and b) and reiteration stability evaluations (c and d) using $TypeA_1$ anomalies. (lower is better).}
    \label{fig:results_reduction}
    \vspace{-2mm}
\end{figure*}

\section{Experimental Results}
\label{sec:experimental_results}

In this section, the experimental results are presented. Thereby, the introduced evaluation framework is used to compare the (i) \textit{fidelity}, (ii) \textit{stability}, and (iii) \textit{robustness} of RESHAPE and baseline XAI methods.

\vspace*{2mm}

\subsection{Explanation Fidelity}

The explanation fidelity of RESHAPE and baseline XAI methods is compared using the three evaluation metrics described in Sec. \ref{sec:explanation_fidelity}.

\subsubsection{Mean Reciprocal Rank ($MRR_r$) Results:} We use the Synthetic Boolean Dataset to evaluate the $MRR_r$ of the different methods. Since the `ground-truth' relationships in the dataset are known, it allows a qualitative assessment of the explanations' attribute rankings. The $MRR_r$ results obtained for the boolean \textit{AND} or \text{OR} dependencies are depicted in Tab. \ref{tab:results_sbd_and}. The results for both dependencies show that the RESHAPE and the A-SHAP methods exhibit a high degree of fidelity. Both methods rank at least one of the attributes of the \textit{AND} or \text{OR} dependency at the top of the explanation. Comparing the methods in terms of ranking independent and dependent variables, A-SHAP identifies the independent variables as the most explanatory. In contrast, the loss-based methods emphasize the dependent variables more.

\begin{table}[htbp]
\centering
\caption[Mean Reciprocal Rank Results]{Synthetic Boolean Dataset $MRR_r$ results (higher is better). Perturbations violate the boolean \textit{AND} and \textit{OR} operator dependencies evident in the dataset.}

\label{tab:results_sbd_and}
\resizebox{0.45\textwidth}{!}{%

	\begin{tabular}{@{}lccc@{}}
	 	 & \multicolumn{3}{c}{\textbf{Boolean \textit{AND} dependency}} \\
		\cmidrule(lr){2-4}
		\textbf{Attributes} &
	  	\begin{tabular}[c]{@{}c@{}}Independent or dependent \\ Attributes $\{a_2, a_3, a_{17}\}$\end{tabular} &
	  	\begin{tabular}[c]{@{}c@{}}Independent\\  Attributes $\{a_2, a_{3}\}$\end{tabular} &
	  	\begin{tabular}[c]{@{}c@{}}Dependent\\  Attribute $\{\Tilde{a}_{17}\}$\end{tabular}  \\ \midrule
	  	
	 	Random & 0.33 $\pm$ \scalebox{0.85}{0.292} & 0.27 $\pm$  \scalebox{0.85}{0.277} & 0.16 $\pm$ \scalebox{0.85}{0.184}     \\ 
 LossSHAP \cite{takeishi2019}  & 0.93 $\pm$ \scalebox{0.85}{0.188} & 0.43 $\pm$ \scalebox{0.85}{0.253} & 0.81 $\pm$ \scalebox{0.85}{0.325} \\
	A-SHAP \cite{antwarg2019} & \textbf{1.00} $\pm$ \textbf{\scalebox{0.85}{0.000}} & \textbf{0.98} $\pm$ \textbf{\scalebox{0.85}{0.122}} & 0.17 $\pm$ \scalebox{0.85}{0.163}   \\
		\midrule
		RESHAPE (ours) & \textbf{1.00} $\pm$ \textbf{\scalebox{0.85}{0.000}} & 0.61 $\pm$ \scalebox{0.85}{0.206} & \textbf{0.86} $\pm$ \textbf{\scalebox{0.85}{0.275}}    \\
		\bottomrule
		\multicolumn{4}{l}{\scalebox{0.8}{*Variance originate from 100 different explained anomalies.}}
		\\ &&&\\

	 	& \multicolumn{3}{c}{\textbf{Boolean \textit{OR} dependency}} \\
		\cmidrule(lr){2-4}
		\textbf{Attributes} &
	  	\begin{tabular}[c]{@{}c@{}}Independent or dependent \\ Attributes $\{a_4, a_5, a_{18}\}$\end{tabular} &
	  	\begin{tabular}[c]{@{}c@{}}Independent\\  Attributes $\{a_4, a_{5}\}$\end{tabular} &
	  	\begin{tabular}[c]{@{}c@{}}Dependent\\  Attribute $\{\Tilde{a}_{18}\}$\end{tabular} \\ \midrule

		Random &  0.33 $\pm$ \scalebox{0.85}{0.296} & 0.26 $\pm$ \scalebox{0.85}{0.262} & 0.17 $\pm$ \scalebox{0.85}{0.213}  \\ 
		LossSHAP \cite{takeishi2019} & 0.93 $\pm$ \scalebox{0.85}{0.183} & 0.49 $\pm$ \scalebox{0.85}{0.236} & 0.79 $\pm$ \scalebox{0.85}{0.345} \\
		A-SHAP \cite{antwarg2019} & \textbf{1.00} $\pm$ \textbf{\scalebox{0.85}{0.000}}  & \textbf{1.00} $\pm$ \textbf{\scalebox{0.85}{0.000}} & 0.20 $\pm$ \scalebox{0.85}{0.032} \\
		\midrule
		RESHAPE (ours) & \textbf{1.00} $\pm$ \textbf{\scalebox{0.85}{0.000}} & 0.64 $\pm$ \scalebox{0.85}{0.222} & \textbf{0.82} $\pm$ \textbf{\scalebox{0.85}{0.296}} \\
		\bottomrule
		\multicolumn{4}{l}{\scalebox{0.8}{*Variance originate from 100 different explained anomalies.}}
	\end{tabular}%
}
\vspace{-0.4cm}
\end{table}

\vspace*{2mm}

\subsubsection{Hits@n Results:} 

We use the Synthetic Accounting and Real-World Payment Dataset to evaluate the $Hits@n$ of the different methods. A high score indicates that an unusual attribute $\Tilde{a}_j^i$ is more often among the $n$ top-ranked attributes of the explanations. The Area Under the Curve (AUC) of the $Hits@n$ over the datasets and XAI methods are shown in Tab. \ref{tab:hits_auc}. The results show that RESHAPE outperforms the other methods in ranking the anomalous attribute values among the top explanatory top attributes. In contrast, the A-SHAP method doesn't exhibit a good performance on all types of anomalous attributes. This originates from the fact that the injected anomalous $TypeA_k$ attributes contain random attribute values. Therefore, the AENN model can't leverage knowledge from other attributes to correctly reconstruct the unusual attribute values. As a result, the A-SHAP explanations fail to explain the $TypeA_k$ anomalies.

\begin{table}[htbp]
\centering
\caption[Area under the $Hits@n$ Curve (AUC) for $TypeA_1$ anomalies]{Area under the $Hits@n$ Curve (AUC) per XAI Methods for $TypeA_1$ anomalies. Variance stems from twenty different anomalies explained and repetition using five AENNs trained using different random seeds (higher is better).}
\label{tab:hits_auc}
\resizebox{0.35\textwidth}{!}{%
\begin{tabular}{@{}lcc@{}}
XAI Method & Accounting Data & Payment Data \\ \midrule
Random & 4.11 $\pm$ \scalebox{0.85}{2.04} & 5.15 $\pm$ \scalebox{0.85}{2.91} \\ 
LossSHAP \cite{takeishi2019} & 5.25 $\pm$ \scalebox{0.85}{2.50} & 6.21 $\pm$ \scalebox{0.85}{2.78} \\
A-SHAP \cite{antwarg2019} & 2.94 $\pm$ \scalebox{0.85}{2.26} & 3.41 $\pm$ \scalebox{0.85}{2.96} \\
\midrule
RESHAPE (ours) & \textbf{6.61} $\pm$ \textbf{\scalebox{0.85}{1.15}} & \textbf{7.37} $\pm$ \textbf{\scalebox{0.85}{2.92}} \\ \bottomrule

\multicolumn{3}{l}{\scalebox{0.8}{*Variance originate from 20 different explained anomalies.}}
\end{tabular}%
}
\vspace{-0.4cm}
\end{table}

\vspace*{2mm}

\subsubsection{Reduction of Anomaly Score:} 

We use the Synthetic Accounting and Real-World Payment Dataset to evaluate the \scalebox{0.85}{$error\%_n$} of the different methods. Furthermore, the objective of the explanation runs is to explain the injected $TypeA_1$ anomalies. The obtained \scalebox{0.85}{$error\%_n$} results obtained for both datasets and XAI methods are shown in Fig. \ref{fig:results_reduction}. Comparing the different methods, RESHAPE and LossSHAP show a comparatively rapid reduction of anomaly scores. The results demonstrate that both methods provide effective explanations for the $TypeA_1$ anomalies in both datasets. At the same time, the A-SHAP exhibits a similar performance to random guessing when explaining the most salient attributes.

\subsection{Explanation Stability}

We use the Synthetic Accounting and Real-World Payment Dataset to evaluate the Stability Index of the top-ranked explanatory attributes by the different methods. The results of the stability evaluation are depicted in Fig. \ref{fig:results_reduction}. It can be observed that RESHAPE and LossSHAP outperform the A-SHAP method on both datasets. Ultimately, RESHAPE results in the lowest Stability Index score of the top-ranked attribute. This indicates that the methods are more certain in terms of which attributes had the highest effect on the reconstruction error of an anomalous JE. 



\vspace*{2mm}

\subsection{Explanation Robustness}

We use the Synthetic Accounting and Real-World Payment Dataset to evaluate the Mean Reciprocal Rank of uninformative attributes ($MRR_u$) score of the different methods. The obtained results over the distinct datasets and XAI methods are shown in Tab. \ref{tab:hits_auc}. For both datasets, the LossSHAP method attributes the least importance to the uninformative attribute, whilst a random ordering of attributes leads to the highest $MRR_u$ being assigned to the uninformative attribute. The RESHAPE method assigns a higher importance to the uninformative attribute than the LossSHAP method but outperforms the A-SHAP method.

\begin{table}[htbp]
\small
\centering
\caption{Results of robustness against noise evaluation: Mean reciprocal rank ($MRR_u$) of noisy attribute for Accounting Data and Transaction Data (lower is better). }
\label{tab:results_noise}
\resizebox{0.35\textwidth}{!}{%
\begin{tabular}{@{}lcc@{}}
XAI Method & Accounting Data & Payment Data \\ \midrule
Random & 0.32 $\pm$ \scalebox{0.85}{0.14} & 0.34 $\pm$ \scalebox{0.85}{0.11} \\
LossSHAP \cite{takeishi2019} & \textbf{0.17} $\pm$ \textbf{\scalebox{0.85}{0.02}} & \textbf{0.14} $\pm$ \textbf{\scalebox{0.85}{0.02}} \\
A-SHAP \cite{antwarg2019} & 0.21 $\pm$ \scalebox{0.85}{0.08} & 0.21 $\pm$ \scalebox{0.85}{0.08}   \\
\midrule
RESHAPE (ours) & 0.18 $\pm$ \scalebox{0.85}{0.03} & 0.17 $\pm$ \scalebox{0.85}{0.04} \\

\bottomrule

\multicolumn{3}{l}{\scalebox{0.8}{*Variance originate from 20 different explained anomalies.}}
\vspace*{4mm}
\end{tabular}
}
\end{table}

\begin{figure}[htbp]
    \center
  \includegraphics[width=0.85\linewidth]{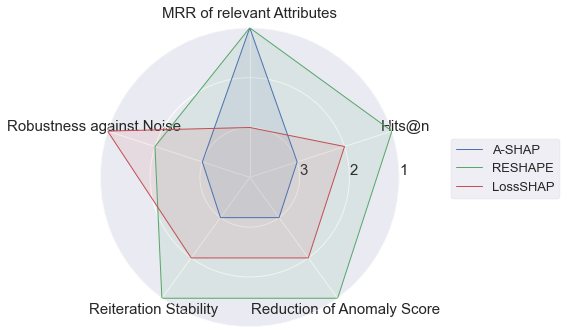}
  \vspace*{4mm}
  \caption{Ranking the performance of compared XAI methods over the set of evaluation metrics.}
  \label{fig:summary}
\end{figure}

\vspace*{2mm}
\noindent Ultimately, the RESHAPE method outperforms the baseline XAI methods in terms of versatility. Figure \ref{fig:summary} summarizes the experiments by ranking each method according to the results obtained for the distinct evaluation metrics.

\section{Summary}
\label{sec:summary}

In this work, we proposed RESHAPE, a novel method to explain the output of AENNs using SHAP. RESHAPE enables the generation of explanations on an attribute level when working with categorical data. Additionally, an evaluation framework for benchmarking XAI methods in a financial audit context is introduced and used to benchmark the currently available methods of explaining AENNs. The assessment against the existing baselines was evaluated on a real-world and two synthetic datasets. The results show that the RESHAPE method is the most versatile methodology in most of the conducted experiments. Therefore, we believe that the RESHAPE method will help to enable the widespread application of complex neural networks in financial audits by providing sufficiently detailed but comprehensible explanations to the auditors, increasing their trust in the opaque models. 

\vspace*{-2mm}

\bibliographystyle{abbrv}
\bibliography{bibliography_thesis}

\end{document}